\documentclass{article}

\PassOptionsToPackage{numbers, compress}{natbib}
\usepackage[preprint]{neurips_data_2023}

\usepackage[utf8]{inputenc} 
\usepackage[T1]{fontenc}    
\usepackage{hyperref}       
\usepackage{url}            
\usepackage{booktabs}       
\usepackage{amsfonts}       
\usepackage{nicefrac}       
\usepackage{microtype}      
\usepackage{listings}
\usepackage{xcolor}
\usepackage{xspace}
\usepackage{amsmath}
\usepackage{comment}
\usepackage{float}
\usepackage[capitalize,noabbrev]{cleveref}
\usepackage{subcaption}
\usepackage[inline]{enumitem}

\usepackage{graphicx, animate}

\definecolor{codegreen}{rgb}{0,0.6,0}
\definecolor{codegray}{rgb}{0.5,0.5,0.5}
\definecolor{codepurple}{rgb}{0.58,0,0.82}
\definecolor{backcolour}{rgb}{0.95,0.95,0.92}

\lstdefinestyle{mystyle}{
    backgroundcolor=\color{backcolour},   
    commentstyle=\color{codegreen},
    keywordstyle=\color{magenta},
    numberstyle=\tiny\color{codegray},
    stringstyle=\color{codepurple},
    basicstyle=\ttfamily\footnotesize,
    breakatwhitespace=false,         
    breaklines=true,                 
    captionpos=b,                    
    keepspaces=true,                 
    numbers=left,                    
    numbersep=5pt,                  
    showspaces=false,                
    showstringspaces=false,
    showtabs=false,                  
    tabsize=2
}
\newcommand{\name}{\texttt{RobocupGym}\xspace}

\hypersetup{
  colorlinks   = true,              
  urlcolor     = blue,              
  linkcolor    = blue,              
  citecolor    = teal                
}
\usepackage{adjustbox} 

\title{\name: A challenging continuous control benchmark in Robocup}

\author{%
Michael Beukman$^{1,2}$\thanks{Correspondence to \texttt{mbeukman@robots.ox.ac.uk}.} \quad Branden Ingram$^{1}$ \quad Geraud Nangue Tasse$^{1}$ \\ \textbf{Benjamin Rosman}$^1$ \quad \textbf{Pravesh Ranchod}$^1$\\\\
$^1$University of the Witwatersrand \quad $^2$University of Oxford
}

\begin{document}

\maketitle

\begin{abstract}
  Reinforcement learning (RL) has progressed substantially over the past decade, with much of this progress being driven by benchmarks.
  Many benchmarks are focused on video or board games, and a large number of robotics benchmarks lack diversity and real-world applicability. In this paper, we aim to simplify the process of applying reinforcement learning in the 3D simulation league of Robocup, a robotic football competition. To this end, we introduce a Robocup-based RL environment based on the open source \texttt{rcssserver3d} soccer server, simple pre-defined tasks, and integration with a popular RL library, Stable Baselines 3.
  Our environment enables the creation of high-dimensional continuous control tasks within a robotics football simulation. In each task, an RL agent controls a simulated Nao robot, and can interact with the ball or other agents.
  We open-source our environment and training code at \url{https://github.com/Michael-Beukman/RobocupGym}.
\end{abstract}

\begin{figure}[H]
    \centering
    \includegraphics[width=\linewidth]{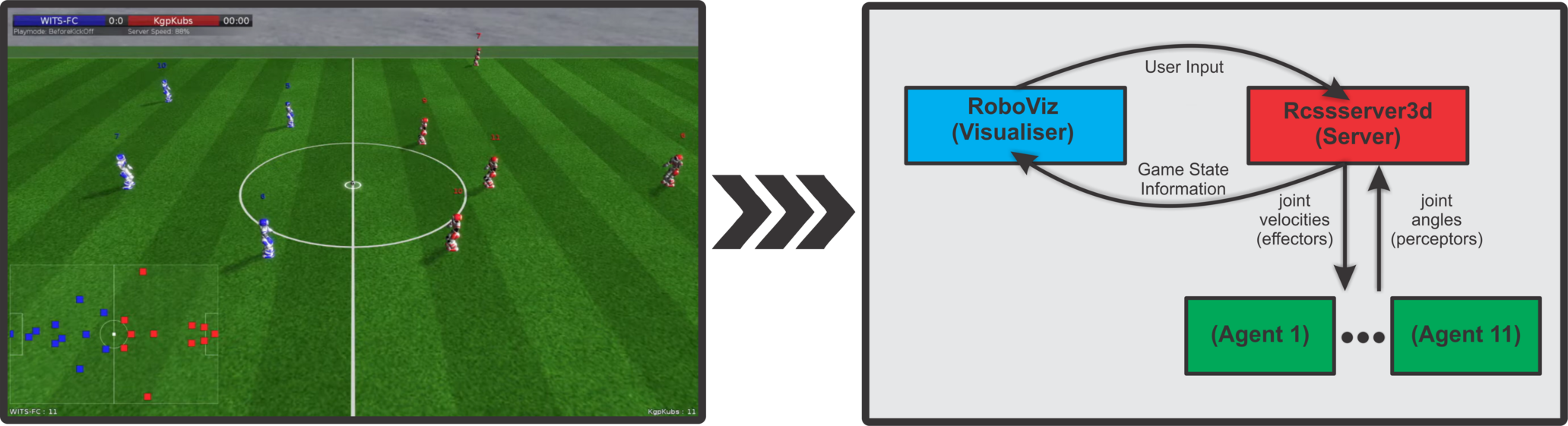}
    \caption{An illustration of the Robocup 3D simulation domain, where two teams of 11 players compete against each other.}
    \label{fig:robocup_domain}
\end{figure}

\section{Introduction}

Reinforcement learning (RL) has achieved remarkable progress in recent years, owing in part to advancements in algorithms~\citep{schulman2015high,schulman2017proximal}, improvements in hardware, and the development and widespread use of benchmark environments~\citep{todorov2012mujoco,bellemare2013arcade,mnih2015human}. These benchmarks provide standardised platforms for assessing the performance of RL algorithms, enabling comparative analysis and tracking of progress within the field. Despite these advancements, there is a clear need for benchmarks that can bridge the gap between simulated environments and real-world applications, particularly in the context of robotics.

Another trend in the field has been to apply RL to more real-world and challenging robotics tasks~\citep{zhu2020Transfer,dulac-arnold2020Empirical,spitznagel2021Deep,kumar2021RMA,gu2024humanoid}.
One notable challenging task is Robocup, a robotic football competition, which has the goal of having a robotic team beat the human football World Cup winners by 2050~\citep{kitano1997robocup}. 
The Robocup competition started in 1997, and has several leagues, ranging from various physical robot morphologies playing football, to a 2D and 3D simulation league and, more recently, other non-football competitions, such as search and rescue and the completion of household tasks. We focus specifically on the 3D simulation football league. This domain is suitable for research on low-level continuous control, and while not a focus of this work, can also be a suitable testbed for multi-agent and hierarchical reinforcement learning.

While RL has previously been applied to this domain~\citep{spitznagel2021Deep,simoesBahiart2022,abreu2023Designing}, the lack of available open-source infrastructure means that it remains difficult for RL practitioners to train low-level continuous control skills in this setting. In particular, other publicly available environments require either the integration with an existing codebase~\citep{simoesBahiart2022,abreu2023Designing} or focus on high-level behaviours~\citep{simoesBahiart2022}.

In our work, we aim to bridge these two areas. To this end, we introduce a convenient reinforcement learning library for the 3D simulation league of Robocup. 
Our library integrates with the Simspark server~\citep{boedecker2008simspark} and allows RL practitioners to easily apply and benchmark RL algorithms on a challenging continuous-control domain. We further integrate with Stable Baselines 3~\citep{sb3}, a commonly used RL library.
The contributions of this work are:

\begin{enumerate}
    \item We present \name, a Python library that allows users to interact with the 3D Robocup simulation.
    \item We provide inbuilt functionality to run RL algorithms in this domain, and provide premade environments.
    \item We provide initial benchmark results of current SoTA RL algorithms on these domains.
\end{enumerate}
\section{\name}
In this section, we describe our environment, \name, how it functions, and how to use it. We conclude this section by describing the technical architecture of \name. See \cref{fig:robocup_domain} for an overview of the Robocup 3D domain.

\subsection{Usage}
\begin{figure} 
    \centering
    
    \begin{subfigure}[t]{0.55\linewidth}
    \captionsetup{width=.0\linewidth}
     \includegraphics[width=\linewidth]{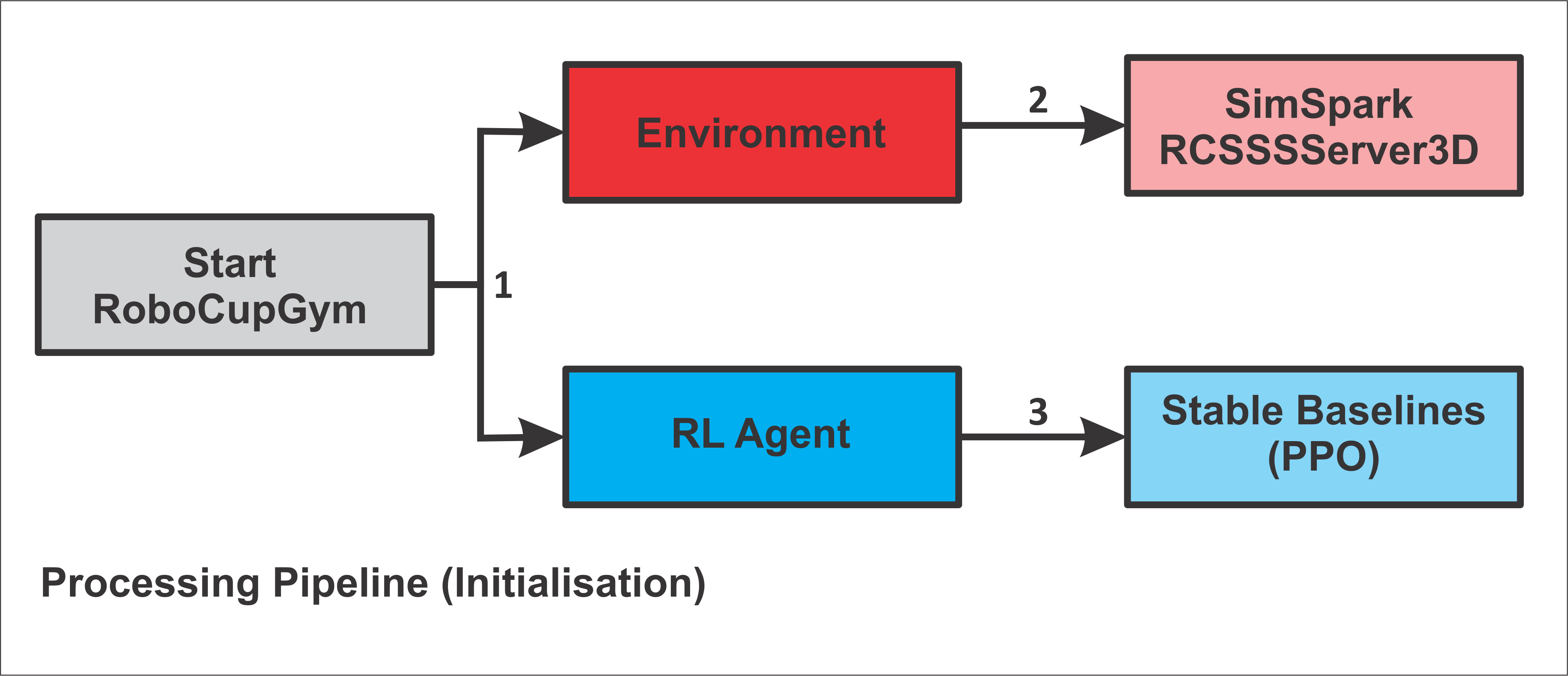}
    \caption{Processing pipeline during initialisation. Step 1: \name creates both the RL environment as well as the RL agent. Step 2: Our environment starts a SimSpark football simulation instance. Step 3: We then initialise an RL agent from Stable Baselines.}
    \label{fig:init}
    \end{subfigure}
    \begin{subfigure}[t]{0.345\linewidth}
    \captionsetup{width=.9\linewidth}
    \includegraphics[width=\linewidth]{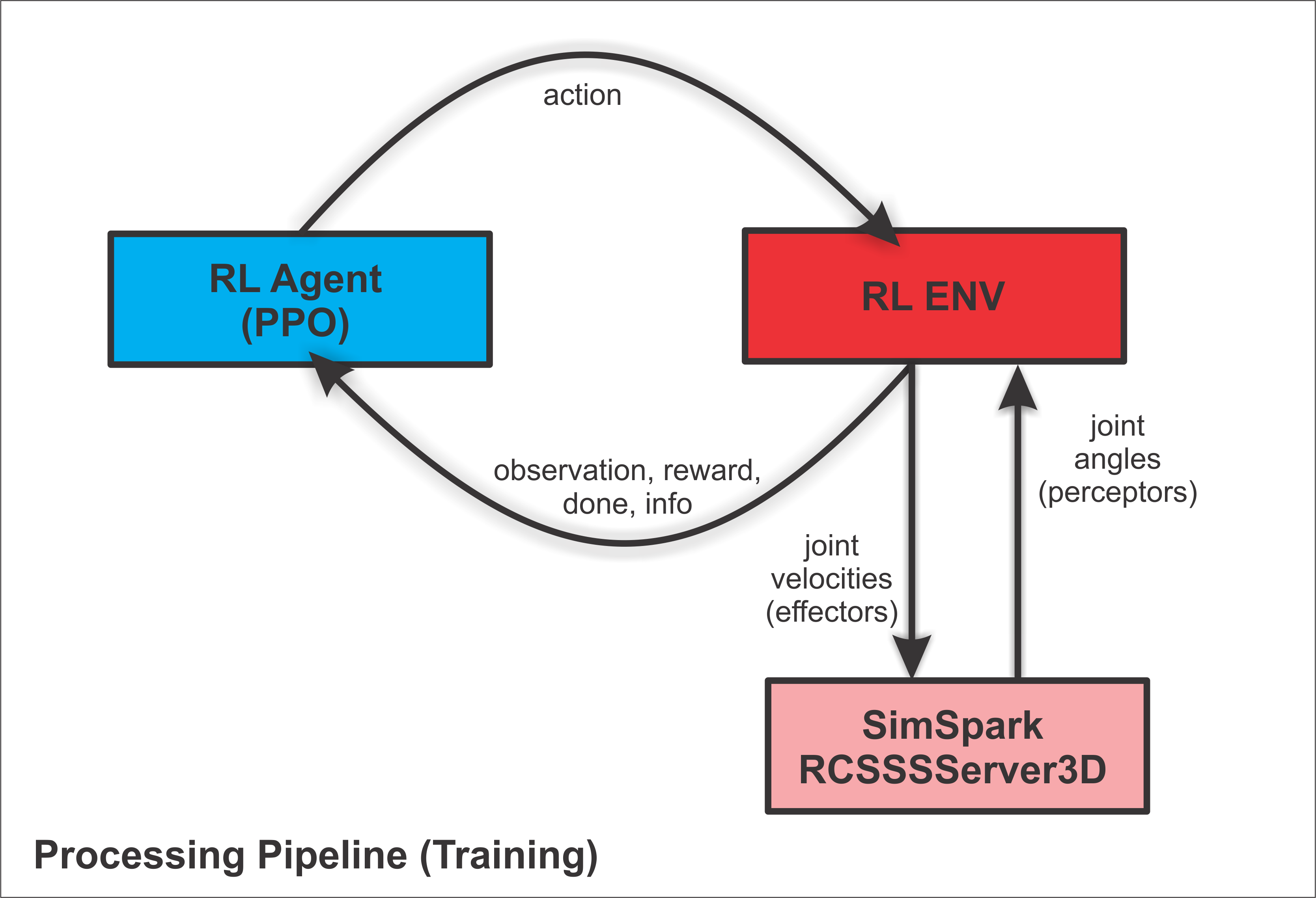}
    \caption{Processing pipeline during training demonstrating the cycle of information between the football server, RL Environment and RL agent.}
    \label{fig:process}
    \end{subfigure}
    \caption{Illustrating the (a) initialisation and (b) processing pipeline of \name.}
\end{figure}

The following code snippet illustrates the usage of \name; the environment can be used as standard in many libraries, such as Stable Baselines. This initialisation process is illustrated in Figure \ref{fig:init}.

\begin{lstlisting}[language=Python]
from robocup_gym.envs.tasks.env_simple_kick import EnvSimpleKick
from stable_baselines3 import PPO

env = EnvSimpleKick()
agent = PPO("MlpPolicy", env)
agent.learn()
\end{lstlisting}

Figure \ref{fig:process} describes the processing pipeline during training. Firstly, The RL environment receives perceptor information from the football server in the form of joint angles and other important information. This information is parsed into an observation which is sent to the RL agent along with the reward, ``is terminal flag'' and additional information fields. The RL agent then returns an action which is passed back to the football server in the form of per-joint velocities. The server processes these effector messages, applies them to the simulation, and the cycle repeats.

\subsection{Existing Tasks}
We provide two variations of a kicking task, one based on the absolute distance the ball is kicked (SimpleKick), and the second based on the ball's velocity after a short waiting time (VelocityKick). 
We choose to start with kicking as it is a core skill in Robocup~\citep{spitznagel2021Deep}. However, our environment is general enough to define various other tasks, allowing users to train skills such as running, goal blocking, etc.

SimpleKick defines the reward function as the x-distance the ball moves, waiting for $N_\text{wait}$ steps after the episode terminates. 
This environment must wait for the ball to stop moving; however, this can lead to most of the time being spent waiting, instead of actually training. 
To alleviate this issue, we propose an alternate reward function as well (VelocityKick): The x-position and x-velocity after a smaller number of timesteps. To incentivise a straight kick, we further penalise y-variation. Finally, in both of these environments, the episode terminates when a maximum number of timesteps has been reached, or the agent has fallen.
\subsection{Adding New Tasks}
We provide functionality to easily create new tasks. 
Each task inherits from the \texttt{BaseEnv} class, and must specify the following aspects:
\begin{description}
    \item[Observation Space] The default observation space consists of the joint positions, joint velocities, relative ball position, force on the robot's foot, and accelerometer and gyroscope sensor readings. This can be customised, however, to include additional aspects, or hide certain information from the agent.
    \item[Action Space] The action space has two parameters, one is which of the 22 joints are controllable, and the second is what each action represents. The three options are: (a) velocity, meaning that the action directly controls the target velocity; (b) target angle, meaning that the action corresponds to a desired angle, and the velocity is chosen accordingly; and (c) desired angle, with a variable maximum speed per joint. By default, all of the joints, except the head and neck, are used, and the action directly determines the velocity.
    \item[Reward Function] The reward function can either be dense or sparse, and can be arbitrarily based on the current simulator state. For instance, the definition of a simple kick task is given below; the environment only overrides the \texttt{\_get\_reward} function, defining the reward to be the distance the ball has moved once the episode ends.
\end{description}
The user can optionally alter other aspects of the MDP, such as the termination condition (by default, the episode terminates after a maximum number of timesteps, or if the agent has fallen).
\begin{lstlisting}[language=Python]
import numpy as np
from robocup_gym.rl.envs.base_env import BaseEnv


class EnvSimpleKick(BaseEnv):
  def _get_reward(self, action: np.ndarray) -> float:
    player = self.python_agent_conn.player
    if self.has_completed:
      current_ball_pos = np.array(player.real_ball_pos)
      start_ball_pos   = np.array(self.env_config.ball_start_pos)
      return np.linalg.norm(current_ball_pos - start_ball_pos)
    return 0

\end{lstlisting}

\subsection{Architecture}
The 3D simulation league in Robocup makes use of the Simspark RCSSserver3D simulator~\citep{boedecker2008simspark}. This is a C/C++ codebase, and uses TCP sockets and S-expressions to communicate with multiple players. 
These players receive information about the simulation, such as information about visible objects, the robot's own joint angles, as well as any communication received from any other agents. The agent then sends a command to the server, which is primarily in the form of a desired angular velocity value for each of the robot's 22 joints.

Each environment in \name spawns an \texttt{rcssserver3d} process, connects to it and instantiates a robot in the environment. Thereafter, the environment allows the agent to perform actions, which are processed and passed to the server. The environment also parses the server messages and converts them into observations for the RL agent.

This architecture allows us to easily create tasks, and treat them as standard RL environments. Indeed, \name follows the standard Gymnasium interface allowing for easy integration into many existing libraries. Furthermore, running multiple environments in parallel is also straightforward to do; this allows for significant wall-clock speedups when running training on a multicore machine.

\section{RL Results}
Here we illustrate results obtained from running standard RL algorithms on \name.
We consider the two existing kick tasks (SimpleKick with 200 wait steps and VelocityKick with 20 wait steps), and use PPO~\citep{schulman2017proximal} and SAC~\citep{haarnoja2018Soft}. Since our environments follow the gym interface~\citep{brockman2016openai}, they can trivially be used with existing RL libraries. We choose to use Stable Baselines 3~\citep{sb3} for all experiments. We use the default SAC hyperparameters, and PPO hyperparameters can be found in \cref{table:ppo_hypers}. Finally, we train agents for 5M timesteps.

\begin{figure}[h]
    \centering
        \includegraphics[width=1\linewidth]{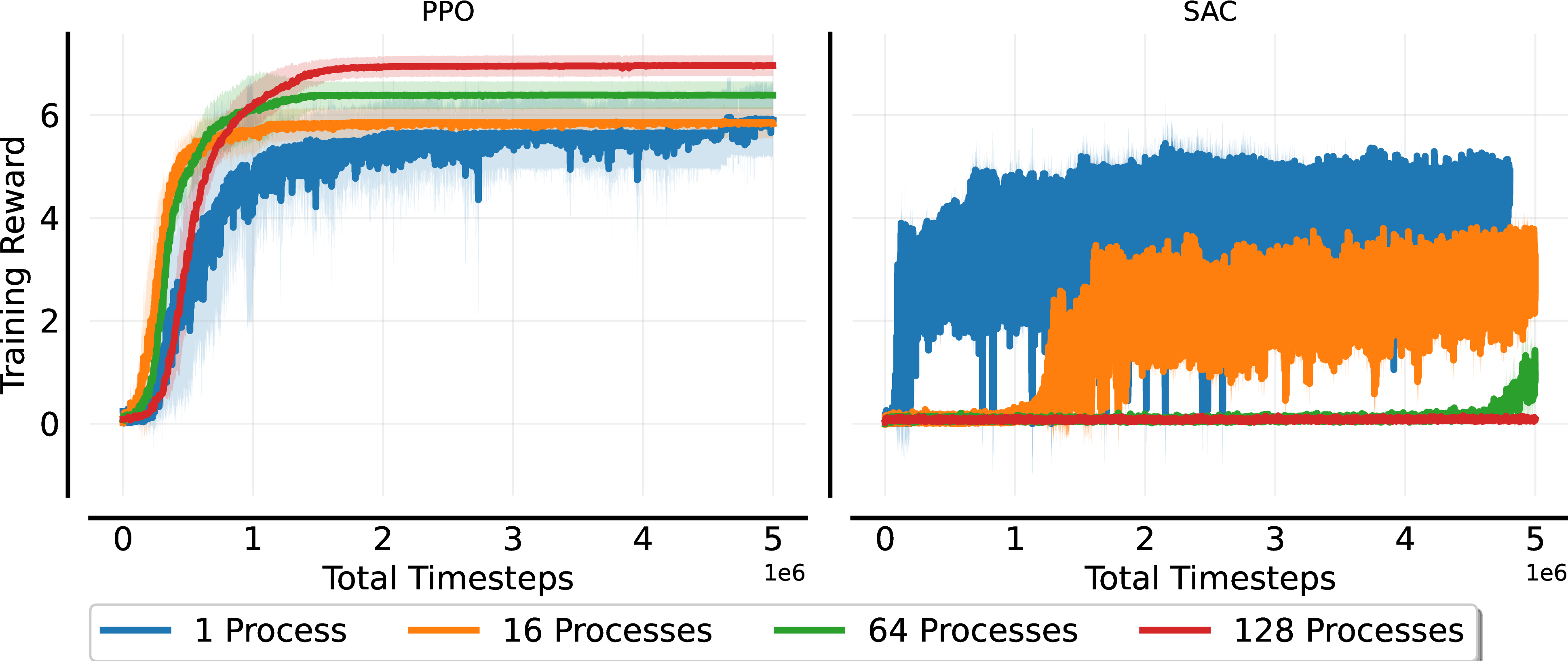}    
        \caption{Comparing PPO and SAC on the VelocityKick environment. PPO outperforms SAC, and scales well as we increase the number of workers.}
    \label{fig:compare_algos:both}
\end{figure}

In \cref{fig:compare_algos:both}, we see that both PPO and SAC are able to learn how to kick the ball, but that PPO performs better overall. Furthermore, when scaling up the number of parallel processes used, PPO maintains or improves performance---given identical hyperparameters---but SAC performs worse.

\begin{figure}[h]
    \includegraphics[width=1\linewidth]{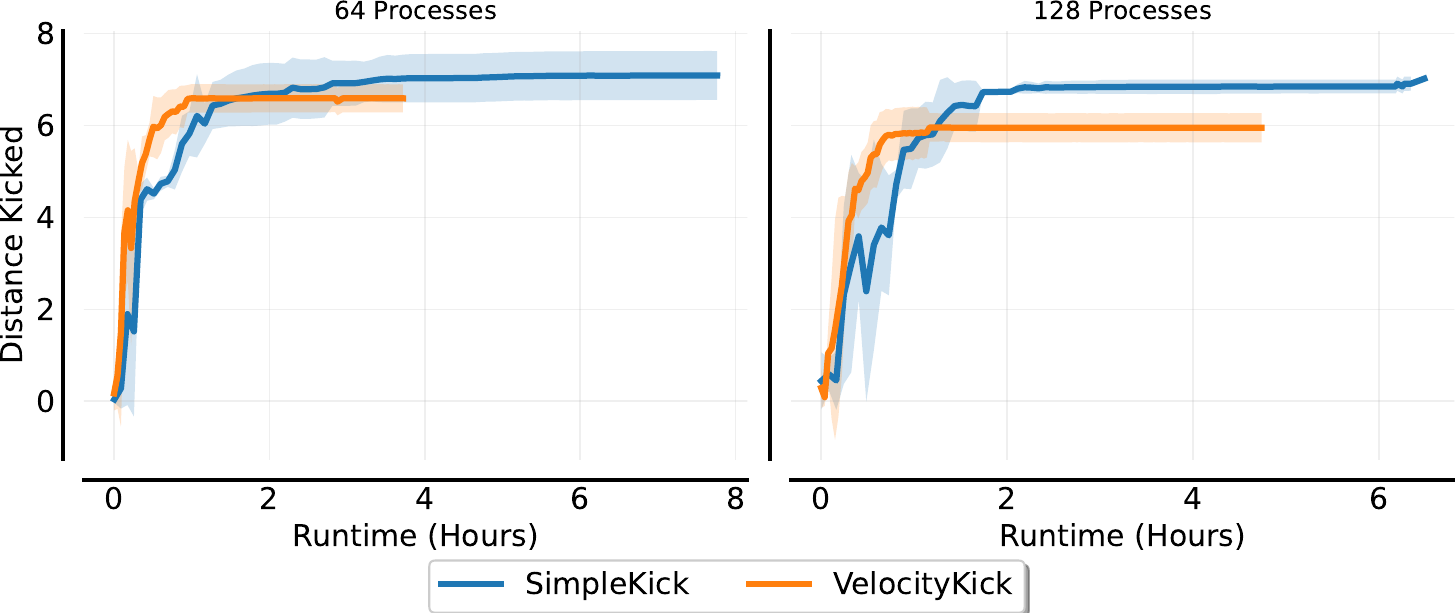}    
    \caption{Comparing the two different kick environments we provide. Both lead to reasonable kicks, but VelocityKick is faster while reaching slightly shorter distances. We use PPO in this plot.}
    \label{fig:compare_envs}
\end{figure}

Comparing the SimpleKick environment to the VelocityKick environment in \cref{fig:compare_envs}, we see that VelocityKick generally results in a slightly shorter kick, but converges much faster than the SimpleKick environment.

\begin{figure}[h]
    \centering
    \includegraphics[width=0.5\linewidth]{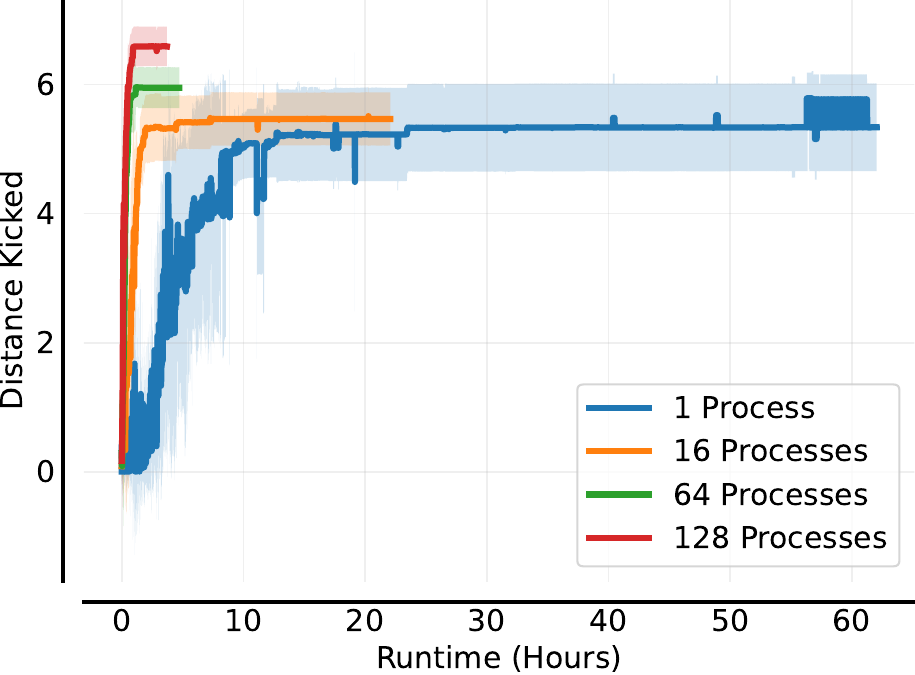}    
    \caption{Runtime for a varying number of processes (for PPO) on the VelocityKick environment. We find a reduction in runtime from around 60 hours with one worker to three hours for 128 workers.}
    \label{fig:compare_runtime_procs}

\end{figure}

Finally, \cref{fig:compare_runtime_procs} illustrates the wall clock time benefit of using multiple processors, allowing experiments to be run over a period of hours instead of days. Furthermore, using more processes improves performance and reduces variance.

\section{Related Work}

Reinforcement learning (RL) has a rich history of benchmarks. Initially, simple and classic tasks were primarily utilised for debugging and proof-of-concept purposes~\citep{michie1968boxes,moore1990efficient,doya2000reinforcement,dejong1994swinging,furuta1978computer,coulom2002reinforcement,schulman2015trust,levine2013guided,sutton2018reinforcement}. In the era of deep RL, Atari games, encapsulated within the Arcade Learning Environment~\citep{bellemare2013arcade}, have become a popular benchmark~\citep{mnih2015human,guo2014deep}. However, with 57 different Atari games, training and evaluating on this benchmark is computationally intensive (e.g., standard practice is to train for 200M frames \textit{per game}~\citep{machado2018revisiting}). Furthermore, all Atari games are singleton environments (i.e., every episode has an identical or very similar configuration), which require limited generalisation to solve~\citep{cobbe2020leveraging}.
To address this, Procgen aims to provide a similar set of tasks but incorporates procedural generation to test generalisation across different levels within the same game~\citep{cobbe2019procgen}. While there is significant diversity in transition dynamics, both benchmarks rely on pixel-based observations and discrete actions. In contrast, \name focuses on tasks more applicable to robotics, featuring continuous state and action spaces.

Beyond these, several robotics-based benchmarks have been developed, most notably MuJoCo~\citep{todorov2012mujoco}. MuJoCo is a general-purpose physics engine with several premade tasks, most of which involve a particular morphology performing a specific task, such as walking in a straight line. The DeepMind Control Suite~\citep{tassa2018deepmind} uses MuJoCo and includes a larger variety of built-in environments. Recently, another environment, Humanoid Gym~\citep{gu2024humanoid}, introduced a framework for training humanoid locomotion policies, focusing on transferring between simulation and the real world. 
Furthermore, several benchmarks aim to test specific aspects of RL, such as safety~\citep{ji2023safety}, meta-learning~\citep{yu2020meta}, and offline RL~\citep{fu2020d4rl}. 
\name could potentially be used to study meta-learning, and the extensive history of Robocup competitions could provide a substantial offline dataset.\footnote{E.g., \url{https://archive.robocup.info/Soccer/Simulation/3D/binaries/Robocup/} contains binaries dating back to 2008.}

Recently, there has been a trend of reimplementing existing environments and creating new ones in Jax~\citep{jax2018github}. These environments enable very fast RL training and rapid experimentation due to hardware acceleration~\citep{brax2021Github,gymnax2022github,nikulin2023xlandMiniGrid,flair2023jaxmarl,matthews2024Craftax}. One downside is that this may shift research towards low sample-efficiency methods, feasible in a hardware-accelerated paradigm (e.g., \citet{matthews2024Craftax} run for 1\textbf{B} timesteps). However, these algorithms may be impractical to run on more traditional simulators or real robots. In contrast, \name uses an established simulator with a long history of use~\citep{boedecker2008simspark,gabel2011progress,macalpine2012ut,macalpine2015ut,macalpine2018ut,abreu2022fc}, and since it is primarily CPU-based, it imposes practical constraints on the number of samples available for training.

\section{Conclusion and Future Work}
This paper introduces \name, a benchmark suite and library designed for training agents in the Robocup 3D simulator. We present two kick-based tasks and demonstrate that current RL agents can quickly learn to perform complex behaviours, such as kicking the ball. Our framework is designed to be easily extendable, allowing new tasks to be added with minimal effort. Useful future work would involve incorporating additional tasks, such as locomotion-based challenges.
Currently, our framework supports only single-agent tasks, but extending it to multi-agent settings would be particularly valuable. Our ultimate goal is to enable the creation of a full Robocup team using exclusively RL. This ambitious aim would involve tackling complex and challenging scenarios, requiring agents to learn both high-level strategies and low-level behaviours. Achieving this would necessitate overcoming many challenges in hierarchical RL, multi-agent RL, and continuous control.
Ultimately, we hope that \name will facilitate more RL research in realistic settings, fostering advancements in the field and contributing to the development of more sophisticated and capable RL agents.
\section*{Acknowledgements}
We thank BahiaRT-Gym for providing inspiration for this project. Discussions with members of the UT Austin Villa, Magma Offenburg and FC Portugal Robocup teams also inspired this project. In particular, thanks to Nico Bohlinger for fruitful discussions and advice.
\bibliographystyle{my_unsrtnat}
\small
\bibliography{reference}

\appendix
\begin{table}[h!]
    \caption{PPO Hyperparameters}
    \label{table:ppo_hypers}
    \begin{center}
        \begin{tabular}{lr}
        \toprule
        \textbf{Parameter}              & PPO  \\
        \midrule
        \textbf{PPO}                    &         \\
        Total Number of Timesteps       & 5M     \\
        $\gamma$                        & 0.99     \\
        $\lambda_{\text{GAE}}$          & 0.95      \\
        PPO number of steps             & 64       \\
        PPO epochs                      & 10         \\
        PPO clip range                  & 0.2       \\
        Adam learning rate              & 0.0001     \\
        entropy coefficient             & 0.0 \\
        observation clipping            & 1.0\\
        \bottomrule 
        \end{tabular}
    \end{center}
\end{table}
\end{document}